\documentclass[12pt]{article}

\usepackage{amssymb}
\usepackage{amsmath}
\usepackage{amscd}
\usepackage{latexsym}
\usepackage{graphicx}

\usepackage{cite}
\usepackage{enumitem}

\newcommand{\be}{\begin{equation}}
\newcommand{\ee}{\end{equation}}

\newcommand{\dlt}{\delta}

\newcommand{\bt}{\beta}

\newcommand{\al}{\alpha}

\newcommand{\cP}{{\cal P}}
\newcommand{\cH}{{\cal H}}

\newcommand{\cA}{{\cal A}}

\newcommand{\rgl}{\rangle}
\newcommand{\lgl}{\langle}

\begin{document}

\begin{center}

{\Large{\bf Quantum Uncertainty in Decision Theory} \\ [5mm]

V.I. Yukalov } \\ [3mm]

{\it Bogolubov Laboratory of Theoretical Physics, \\
Joint Institute for Nuclear Research, Dubna 141980, Russia \\ [2mm]
and \\ [2mm]
Instituto de Fisica de S\~ao Carlos, Universidade de S\~ao Paulo, \\
CP 369,  S\~ao Carlos 13560-970, S\~ao Paulo, Brazil  \\ [5mm]
E-mail: yukalov@theor.jinr.ru} 
\end{center}

\vskip 5cm

\begin{abstract}
An approach is presented treating decision theory as a probabilistic theory based 
on quantum techniques. Accurate definitions are given and thorough analysis is 
accomplished for the quantum probabilities describing the choice between separate 
alternatives, sequential alternatives characterizing conditional quantum probabilities, 
and behavioral quantum probabilities taking into account rational-irrational duality 
of decision making. The comparison between quantum and classical probabilities is 
explained. The analysis demonstrates that quantum probabilities serve as an essentially 
more powerful tool of characterizing various decision-making situations including the 
influence of psychological behavioral effects. 
\end{abstract}

\vskip 1cm


\newpage

\section{Introduction}

Decision theory is an interdisciplinary topic widely used in various applications,
such as economics, statistics, finances, data analysis, psychology, biology, politics, 
social sciences, philosophy, in computer and artificial intelligence studies. For 
example, one has to decide choosing between several portfolios of assets. The basis 
of decision theory is expected utility theory \cite{Neumann_Morgenstern_1} that is 
a deterministic normative theory prescribing that with probability one the optimal 
choice is associated with the largest expected utility. There exist as well stochastic 
decision theories \cite{Berger_2,Raiffa_Schlaifer_3}, which nevertheless are based 
on a deterministic approach decorated by superimposed randomness of utility. 

However the overwhelming majority of empirical studies demonstrates that the choice 
of decision makers is not deterministic but in principle probabilistic. It is never 
happens that among a given group of people all without exception would make the 
identical choice prescribed by the standard deterministic utility theory. There always 
exist fractions of subjects preferring different alternatives. That is, there always 
exists a distribution of decisions over the set of the given alternatives. Moreover, 
as has been recently summarized by Woodford \cite{Woodford_4}, psychological and 
neurological studies persuasively demonstrate that even a single decision maker in 
a single choice acts probabilistically because of the noisy functioning of the brain. 
This tells us that the correct description of decision making has to be probabilistic. 

Two types of probabilities are known, classical and quantum. The natural question is:
Which of these two types is more appropriate for describing human decision making? To 
convincely answer this question, it is necessary to thoroughly compare what situations 
in decision making can be treated by classical and quantum probabilities. That of them 
having the wider region of applicability would be more general, hence preferable, 
although the other one also can be employed in its domain of validity.

The basics of classical probability theory are well known \cite{Kolmogorov_5}. 
Quantum probability, as applied to physical measurements, also has been described 
in several books \cite{Neumann_6}. Here we give the exposition of quantum probability 
in the language of decision theory, keeping in mind a decision maker performing a 
choice between several alternatives. We start from the notion of the probability 
of a single choice, then consider sequential choices among alternatives and compare 
these with classical probabilities.
 
It is important to stress that quantum probabilities can be generalized to the form 
that allows us to take into account behavioral effects and the dual nature of human 
decision making, comprising its rational cognitive side as well as irrational emotional 
feelings. This generalization also is described.

\section{Probability of separate alternatives}

A decision problem is assumed to consist in choosing one of the alternatives from 
the set $\{A_n:\; n = 1,2,\ldots,N_A\}$ of alternatives. Each alternative is associated 
with a vector $|A_n\rangle$ from a Hilbert space $\mathcal{H}_A$. Here and in what 
follows, we employ the Dirac bracket notation \cite{Dirac_7}. It is possible to accept 
that the set of the vectors $|A_n\rangle$ forms a basis of $\mathcal{H}_A$, although, 
for generality, we can accept any basis whose span composes the space $\mathcal{H}_A$. 
The vectors $|A_n\rangle$ are orthonormalized, such that
\be
\label{1}
\lgl A_m \; | \; A_n \rgl = \dlt_{mn} \qquad ( n = 1,2, \ldots, N_A) \; .
\ee

Since we are planning to develop a probabilistic approach, we need a statistical 
object in order to introduce later quantum probability. The statistics of the 
problem at time $t$ is represented by a statistical operator $\hat{\rho}(t)$, which is 
a semipositive trace-class operator on $\cH_A$, briefly called the state. The pair 
$\{\cH_A,\hat{\rho}(t)\}$ forms a {\it quantum statistical ensemble}. The evolution 
of the state in time, from an initial state $\rho(0)$, is given by the relation
\be
\label{2}
  \hat\rho(t) = \hat U(t,0) \;   \hat\rho(0) \; \hat U^+(t,0)
\ee
through the action of a unitary evolution operator $\hat{U}(t,0)$.

The set of alternatives is usually associated with the eigenvectors of operators 
of local observables acting on the Hilbert space $\cH_A$. The collection of these 
operators $\{\hat{A}\}$ composes a von Neumann algebra $\cA(\cH_A)$ which is a 
norm-closed algebra of bounded self-adjoint operators, containing the identity 
operator, on the Hilbert space $\cH_A$. The triple 
$$
\{ \cH_A , \; \hat\rho(t) , \; \cA(\cH_A) \}   
$$
is termed a {\it quantum statistical system}.    

To each alternative, one puts into correspondence a projector
\be
\label{3}
\hat P(A_n) = |\; A_n \rgl \lgl A_n \; | \;   .
\ee
In quantum measurements, there can occur the so-called degenerate states with 
projectors given by a sum of projectors. However in decision theory the degenerate 
states need to be specified, otherwise they do not have much sense. Therefore in 
decision theory we shall deal with the standard projectors (\ref{3}). The collection 
of these projectors constitutes a projector-valued measure 
\be
\label{4}
 \cP(A) = \{ \hat P(A_n): \; n = 1,2,\ldots,N_A \} \; .
\ee
And the triple
\be
\label{5}
\{ \cH_A , \; \hat\rho(t) , \; \cP(A) \}  
\ee
is a {\it quantum probability space}.

The probability of choosing at the moment of time $t$ an alternative $A_n$ is
\be
\label{6}
 p(A_n,t) = {\rm Tr}\; \hat\rho(t) \; \hat P(A_n) \; ,
\ee  
with the trace over $\mathcal{H}_A$. The probability is normalized, so that
\be
\label{7}
 \sum_n \; p(A_n,t) = 1 \; , \qquad 0 \; \leq \; p(A_n,t) \; \leq \; 1 \; .
\ee
Unfolding the probability (\ref{6}), we get
\be
\label{8}
  p(A_n,t) = \lgl A_n \; | \; \hat\rho(t) \; | \; A_n \rgl \;  .
\ee
Note that this result does not depend on the used basis, if one takes into account 
that for any basis $\{|j\rangle\}$, one has
$$
 \sum_j \; |\; j \rgl \lgl j \; | = \hat 1 \; .
$$
 
Similarly, for any problem represented by an operator $\hat{B}\in\cA(\cH_A)$, one has 
a set of vectors $\{|B_k\rangle\}$ that are ortonormalized,
\be
\label{9}
\lgl B_k \; | \; B_p \rgl = \dlt_{kp} \qquad ( k = 1,2,\ldots,N_B )
\ee
and define the corresponding projectors
\be
\label{10}
\hat P(B_k) = |\; B_k \rgl \lgl B_k \; | \;  .
\ee
The family of these projectors composes the operator valued measure
\be
\label{11}
\cP(B) = \{ \hat P(B_k): \; k = 1,2,\ldots,N_B \} \; .
\ee
The corresponding quantum probability space is
\be
\label{12}
  \{\cH_A , \; \hat\rho(t) , \; \cP(B) \} \; .
\ee
The probability of choosing at time $t$ an alternative $B_k$ is
\be
\label{13}
p(B_k,t) = {\rm Tr} \; \hat\rho(t) \; \hat P(B_k)   
\ee
satisfying the normalization condition
\be
\label{14}
 \sum_k p(B_k,t) = 1 \; , \qquad 0 \; \leq \; p(B_k,t) \; \leq \; 1 \; .
\ee

\section{Probability of sequential choices}

Considering sequential choices, it is necessary to accurately take into account the 
temporal evolution of probabilities \cite{Yukalov_8}. The probability of an alternative 
$A_n$, defined in Eq. (\ref{6}), is the a priori expected probability of choosing this 
alternative. Suppose at the moment $t_0$ a decision has been made and the alternative 
$A_n$ has actually been chosen. Hence a priori form (\ref{6}) is valid only before this 
time up to the time $t_0 - 0$, when 
\be
\label{15}
p(A_n,t_0-0) = {\rm Tr}\; \hat\rho(t_0-0) \; \hat P(A_n) \; .
\ee
If at the time $t_0$ the alternative $A_n$ has certainly been chosen, this means that 
the a posteriori probability of choosing $A_n$ becomes
\be
\label{16}
 p(A_n,t_0+0) = 1 \; .
\ee
One tells that at the moment $t_0$ the state reduction has happened,
\be
\label{17}
 \hat\rho(t_0-0) \longmapsto  \hat\rho(A_n,t_0+0)\; ,   
\ee
leading to the probability update
\be
\label{18}
 p(A_n,t_0-0) \longmapsto p(A_n,t_0+0) \; .
\ee

Condition (\ref{16}) implies
\be
\label{19}
  p(A_n,t_0+0) = {\rm Tr}\; \hat\rho(A_n,t_0+0) \; \hat P(A_n) = 1 \; .
\ee
The solution to this equation, describing the state reduction (\ref{17}), reads as
\be
\label{20}
\hat\rho(A_n,t_0+0) = 
\frac{\hat P(A_n)\hat\rho(t_0-0)\hat P(A_n)}{{\rm Tr}\hat\rho(t_0-0)\hat P(A_n)} \; ,
\ee
which is the von Neumann-L\"{u}ders state \cite{Neumann_6,Luders_9}. This form serves 
as a new initial condition for the state evolution
\be
\label{21}
 \hat\rho(A_n,t) = \hat U(t,t_0) \; \hat\rho(A_n,t_0+0) \; \hat U^+(t,t_0)  
\ee
after the time $t_0$.

Aiming at making a choice among the set of alternatives $\{B_k\}$, we deal with the 
quantum probability space
\be
\label{22}
 \{ \cH_A , \; \hat\rho(A_n,t) , \; \cP(B) \} \; .
\ee
The related probability plays the role of the {\it quantum conditional probability}
\be
\label{23}
p(B_k,t|A_n,t_0) = {\rm Tr}\; \hat\rho(A_n,t) \; \hat P(B_k)
\ee
of choosing an alternative $B_k$ at a time $t$, after the alternative $A_n$ has certainly 
been chosen at the time $t_0$.   

Introducing the notation
\be
\label{24}
p(B_k,t,A_n,t_0) \equiv {\rm Tr}\;\hat U(t,t_0) \; \hat P(A_n)\;  
\hat\rho(t_0-0) \; \hat P(A_n) \; \hat U^+(t,t_0) \; \hat P(B_k)
\ee
allows us to represent the conditional probability (\ref{23}) in the form
\be
\label{25}
 p(B_k,t|A_n,t_0) = \frac{p(B_k,t,A_n,t_0)}{p(A_n,t_0-0)} \; .
\ee
This relation is similar to the classical relation between conditional and joint 
probabilities, which justifies the admissibility of naming the probability (\ref{24})
{\it quantum joint probability} of choosing an alternative $A_n$ at the time $t_0$ 
and an alternative $B_k$ at the time $t > t_0$. The difference with the classical 
probabilities is that now the choices are made at different times, but not 
simultaneously. 

An important particular case is when the second choice is made immediately after 
the first one \cite{Goodman_10}. Then the evolution operator reduces to the identity,
\be
\label{26}
 \hat U(t_0+0,t_0) = \hat 1 \; .
\ee
This simplifies the conditional probability
\be
\label{27}
p(B_k,t_0+0|A_n,t_0) = {\rm Tr}\; \hat\rho(A_n,t_0+0) \; \hat P(B_k)
\ee
and the joint probability
\be
\label{28}
 p(B_k,t_0+0,A_n,t_0) = {\rm Tr}\; \hat P(A_n) \; \hat\rho(t_0-0) \; 
\hat P(A_n) \; \hat P(B_k) \; .
\ee   
The latter is analogous to the Wigner probability \cite{Wigner_11}. The relation 
(\ref{25}) now reads as
\be
\label{29}
 p(B_k,t_0+0|A_n,t_0) = \frac{p(B_k,t_0+0,A_n,t_0)}{p(A_n,t_0-0)} \; .
\ee
Accomplishing the trace operation in the joint probability (\ref{28}) yields
\be
\label{30}
p(B_k,t_0+0,A_n,t_0) = | \; \lgl B_k \; | \; A_n \rgl \; |^2 \; p(A_n,t_0-0) \;  .
\ee
Hence the conditional probability (\ref{29}) becomes
\be
\label{31}
  p(B_k,t_0+0|A_n,t_0) = | \; \lgl B_k \; | \; A_n \rgl \; |^2 \;  .
\ee

\section{Symmetry properties of probabilities}

It is important to study the symmetry properties of the quantum probabilities when the 
choice order reverses, that is, first one chooses an alternative $B_k$ at the time $t_0$
and then one considers the probability of choosing an alternative $A_n$ at the time $t$.
The symmetry properties should be compared with those of classical probabilities. Not to
confuse the latter with the quantum probabilities, denoted by the letter $p$, we shall 
denote the classical probabilities by the letter $f$. Thus the classical conditional 
probability of two events, $A_n$ and $B_k$, is
\be
\label{32}
 f(B_k|A_n) = \frac{f(B_kA_n)}{f(A_n)} \;  ,
\ee
where $f(B_kA_n)$ is the classical joint probability, which is symmetric:
\be
\label{33}
f(A_nB_k) = f(B_kA_n) \;  ,
\ee
while the conditional probability is not,
\be
\label{34}
 f(A_n|B_k) \neq  f(B_k|A_n) \; .
\ee
   
For the quantum probabilities with the reversed order, acting as in the previous 
section, we obtain the conditional probability
\be
\label{35}
 p(A_n,t|B_k,t_0) = {\rm Tr}\; \hat\rho(B_k,t) \;\hat P(A_n) \; ,
\ee
with the state
\be
\label{36}
 \hat\rho(B_k,t) = \hat U(t,t_0) \; \hat\rho(B_k,t_0+0) \; \hat U^+(t,t_0) \; ,
\ee
where
\be
\label{37}
\hat\rho(B_k,t_0+0) = \frac{\hat P(B_k)\hat\rho(t_0-0)\hat P(B_k)}
{{\rm Tr}\hat\rho(t_0-0)\hat P(B_k)} \; .
\ee
Introducing the notation of the joint probability
\be
\label{38}
 p(A_n,t,B_k,t_0) \equiv {\rm Tr}\; \hat U(t,t_0) \; \hat P(B_k)
\hat\rho(t_0-0) \; \hat P(B_k) \; \hat U^+(t,t_0) \; \hat P(A_n) \; ,
\ee
results in the relation
\be
\label{39}
 p(A_n,t|B_k,t_0) = \frac{ p(A_n,t,B_k,t_0)}{p(B_k,t_0-0)} \;  .
\ee

For different times $t$ and $t_0$, neither conditional nor joint quantum probabilities
are symmetric:
$$
 p(A_n,t|B_k,t_0) \neq  p(B_k,t|A_n,t_0) \; ,
$$
\be
\label{40}
 p(A_n,t,B_k,t_0) \neq  p(B_k,t,A_n,t_0)  \qquad ( t > t_0 ) \; .
\ee

The quantum and classical probabilities satisfy the same normalization conditions, 
such as for the conditional probability
\be
\label{41}
\sum_k p(B_k,t|A_n,t_0) = \sum_n p(A_n,t|B_k,t_0) = 1
\ee  
and for the joint probability
$$
\sum_k p(B_k,t,A_n,t_0) = p(A_n,t_0-0) \; ,
$$
\be
\label{42}
 \sum_n p(A_n,t,B_k,t_0) = p(B_k,t_0-0) \; .
\ee
Then the normalization condition follows:
\be
\label{43}
\sum_{nk}  p(B_k,t,A_n,t_0) = \sum_{nk}  p(A_n,t,B_k,t_0) = 1 \; ,
\ee 
from which  
\be
\label{44}
\sum_{nk} [\; p(B_k,t,A_n,t_0) -  p(A_n,t,B_k,t_0) \; ] = 0 \; .
\ee

In the case when in the second choice at the time $t_0 + 0$ one estimates the 
probability of an alternative $A_n$ immediately after the first choice at the time 
$t_0$ has resulted in an alternative $B_k$, similarly to the previous section, we 
find the joint probability
\be
\label{45}
p(A_n,t_0+0,B_k,t_0) = | \lgl A_n \; | \; B_k \rgl |^2 \; p(B_k,t_0-0)
\ee
and the conditional probability
\be
\label{46}
p(A_n,t_0+0|B_k,t_0) = | \lgl A_n \; | \; B_k \rgl |^2  \; .
\ee
Therefore the conditional probability is symmetric, while the joint probability 
is not:
$$
p(A_n,t_0+0|B_k,t_0) = p(B_k,t_0+0|A_n,t_0) \; ,
$$
\be
\label{47}
p(A_n,t_0+0,B_k,t_0) \neq p(B_k,t_0+0,A_n,t_0) \qquad (t = t_0 +0 ) \; ,
\ee
which is contrary to the classical case (\ref{33}) and (\ref{34}). 

If the projectors $\hat{P}(A_n)$ and $\hat{P}(B_k)$ commute, then the joint 
probability (\ref{28}) becomes symmetric:
\be
\label{48}
 p(B_k,t_0+0,A_n,t_0) = 
{\rm Tr} \; \hat\rho(t_0-0) \; \hat P(B_k) \hat P(A_n) =
p(A_n,t_0+0,B_k,t_0) \; ,
\ee
provided that $[\hat{P}(A_n), \hat{P}(B_k)] = 0$. Taking into account the form of 
the joint probability (\ref{45}), we get the equality
\be
\label{49}
 p(B_k,t_0-0) = p(A_n,t_0-0) \qquad 
\left(\; \left[\; \hat P(A_n) , \; \hat P(B_k) \; \right] = 0 \right) \;  .
\ee
Thus in that case both the joint and conditional probabilities are symmetric, which 
contradicts the asymmetry of the classical conditional probability. 

If the repeated choice is made among the same set of alternatives, say $\{A_n\}$, 
that is when $B_k=A_k$, we obtain
\be
\label{50}
 p(A_k,t_0+0|A_n,t_0) = \dlt_{nk} \; .
\ee
This equation represents the principle of the {\it choice reproducibility}, 
according to which, when the choice, among the same set of alternatives, is made 
twice, immediately one after another, the second choice reproduces the first one. 
This sounds reasonable for decision making. Really, when a decision maker accomplishes 
a choice immediately after another one, there is no time for deliberation, hence this 
decision maker just should repeat the previous choice \cite{Yukalov_12}.

\section{Duality in decision making}

Human decision making is known to be of dual nature, including the rational (slow, 
cognitive, conscious, objective) evaluation of alternatives and their irrational 
(fast, emotional, subconscious, subjective) appreciation 
\cite{Sun_13,Paivio_14,Stanovich_15,Kahneman_16}. This feature of decision making 
that can be called {\it rational-irrational duality}, or {\it cognition-emotion 
duality}, or {\it objective-subjective duality}, can be effectively described in 
the language of quantum theory that also possesses a dual nature comprising the 
so-called particle-wave duality. To take into account the dual nature of decision 
making, the quantum decision theory has been advanced 
\cite{Yukalov_17,Yukalov_18,Yukalov_19,Yukalov_20,Yukalov_21,Yukalov_22}. In the 
frame of this theory, quantum probability, taking account of emotional behavioral 
effects, becomes behavioral probability \cite{Yukalov_23}. Below, we briefly delineate 
quantum decision theory following the recent papers \cite{Yukalov_8,Yukalov_12}.
  
The space of alternatives $\mathcal{H}_A$ is composed of the state vectors 
characterizing the rational representation of these alternatives whose probabilities 
can be rationally and objectively evaluated. Since there also exist subjective 
emotional feelings, for taking them into account, the space of the state vectors 
has to be extended by including the {\it subject space}
\be
\label{51}
\cH_S = {\rm span}\; \{ |\; \al\; \rgl \}
\ee
formed by the vector representations $|\alpha \rangle$ of all admissible elementary 
feelings. These vectors $|\alpha \rangle$ form an orthonormal basis, 
$$
\lgl \; \al \; | \; \bt \; \rgl = \dlt_{\al\bt} \;   .
$$ 
Thus, the total {\it decision space} is the tensor product 
\be
\label{52}
 \cH = \cH_A \; \bigotimes \; \cH_S  \; .
\ee
 
The statistical state $\hat{\rho}(t)$ now acts on the decision space (\ref{52}) 
where it evolves as
\be
\label{53}
\hat\rho(t) = \hat U(t,0) \; \hat\rho(0) \; \hat U^+(t,0) \; .
\ee
Respectively, the quantum statistical ensemble is
\be
\label{54}
\left\{ \cH = \cH_A \; \bigotimes \; \cH_S , \; \hat\rho(t) \right\} \;  .
\ee

Each alternative $A_n$ is accompanied by a related set of emotions $x_n$ that is 
represented in the subject space by an emotion vector $|x_n\rangle\in\cH_S$, which 
can be written as an expansion
\be
\label{55}
 |\; x_n \; \rgl = \sum_\al \; b_{n\al} \; | \; \al \; \rgl \; .
\ee    
Strictly speaking, emotions are contextual and are subject to variations, which 
means that the coefficients $b_{n \alpha}$ can vary and, generally, fluctuate with 
time depending on the state of a decision maker and the corresponding surrounding.    

The emotion vectors can be normalized,
\be
\label{56}
 \lgl \;x_n \; |\; x_n \; \rgl = \sum_\al \; | \; b_{n\al}\;|^2 = 1\; ,
\ee
but they are not necessarily orthogonal, so that 
\be
\label{57}
\lgl \;x_m \; |\; x_n \; \rgl = \sum_\al \;  b^*_{m\al} b_{n\al} 
\ee
is not compulsorily a Kronecker delta.  
           
An emotion operator
\be
\label{58}
\hat P(x_n) = | \; x_n \; \rgl \; \lgl x_n \; | 
\ee
is idempotent,
\be
\label{59}
 [\; \hat P(x_n)\;]^2  = \hat P(x_n) \; ,
\ee
but different operators are not orthogonal, since
\be
\label{60}
 \hat P(x_m) \; \hat P(x_n) = \lgl \;x_m \; |\; x_n \; \rgl   
|\; x_m \; \rgl \lgl \;x_n \; | \; .
\ee
The emotion operators of elementary feelings $|\alpha\rangle$, forming a complete 
orthonormal basis in the space (\ref{51}), sum to one
$$
\sum_\al \hat P(\al) = \sum_\al |\; \al \;\rgl \lgl \; \al \; | = \hat 1 \; ,
$$
but the emotion operators (\ref{58}) do not necessarily sum to one, giving
$$
 \sum_n \hat P(x_n) = \sum_n \; \sum_{\al\bt} \; b_{n\al} b^*_{n\bt}
|\; \al \;\rgl \lgl \; \bt \; | \; ,
$$
from where
$$
\lgl \; \al \; | \; \sum_n \hat P(x_n) \; | \; \bt \; \rgl =
\sum_n  b_{n\al} b^*_{n\bt} \;  .
$$
The emotion vectors $|x_n\rangle$ do not form a basis, hence the emotion operators 
(\ref{58}) do not have to sum to one. The projector (\ref{58}) projects onto the 
subspace of feelings associated with the alternative $A_n$. 

The pair of an alternative $A_n$ and the set of the related emotions $x_n$ composes 
a prospect $A_n x_n$ whose representation in the decision space (\ref{52}) is given 
by the vector
\be
\label{61}
 | \; A_n x_n \; \rgl = | \; A_n \; \rgl \; \bigotimes \; | \; x_n \; \rgl =
\sum_\al \; b_{n\al} \; | \; A_n \al\; \rgl \; .
\ee
These vectors are orthonormalized,
$$
\lgl \;x_m A_m \; |\; A_n x_n \; \rgl = \dlt_{mn} \;  .
$$
The prospect projector
\be
\label{62}
\hat P(A_n x_n) = |\; A_n x_n \; \rgl \lgl \;x_n A_n \; | = 
\hat P(A_n) \; \bigotimes \; \hat P(x_n)
\ee
is idempotent,
\be
\label{63}
 [\; \hat P(A_n x_n) \; ]^2 = \hat P(A_n x_n) \;  .
\ee
The projectors of different prospects are orthogonal,
\be
\label{64}
 \hat P(A_m x_m) \; \hat P(A_n x_n) =  \dlt_{mn} \hat P(A_n x_n)
\ee
and commute with each other,
$$
[\; \hat P(A_m x_m), \; \hat P(A_n x_n) \; ] = 0 \; .
$$

The vectors $|A_n \alpha\rangle$ generate a complete basis in the decision space 
(\ref{52}), because of which
\be
\label{65}
 \sum_{n\al} \; \hat P(A_n\al) = \sum_{n\al} \; 
|\; A_n\al \; \rgl \lgl \; \al A_n \; | = \hat 1 \; .
\ee
But the prospect projectors (\ref{62}) on subspaces do not necessarily sum to one,
$$
\sum_n \; \hat P(A_n x_n) = \sum_n \; \sum_{\al\bt} \; b_{n\al} b_{n\bt}^* \;
|\; A_n \al \; \rgl \lgl \; \bt A_n \; | \; ,
$$
as far as
$$
\lgl \; \al A_m \; | \; \sum_n \; \hat P(A_n x_n) \; |\; A_n\bt \; \rgl =
\dlt_{mn} b_{n\al}^* b_{n\bt} \;   .
$$
However, it is admissible to require that the prospect projectors would sum to one 
on average, so that
\be
\label{66}
 {\rm Tr}\; \hat\rho(t) \; \sum_n \; \hat P(A_n x_n) = 1 \;  ,
\ee
which is equivalent to the condition
\be
\label{67}
\sum_n \; \sum_{\al\bt} \; b^*_{n\al} b_{n\bt} \;  
\lgl \; \al A_n \; | \; \hat\rho(t) \; | \; A_n\bt \; \rgl = 1 \; .
\ee
The trace operation in Eq. (\ref{66}) and below is over the total decision 
space (\ref{52}). 
 
The projection-valued measure on the space (\ref{52}) is
\be
\label{68}
 \cP(A x) = \{ \hat P(A_n x_n): \; n = 1,2,\ldots,N_A \} \;  ,
\ee
so that the quantum probability space is
\be
\label{69}
\{ \cH , \; \hat\rho(t) , \; \cP(A x) \}  \;  .
\ee
The prospect probability reads as
\be
\label{70}
p(A_n x_n,t) = {\rm Tr}\; \hat\rho(t) \; \hat P(A_n x_n)   
\ee
and satisfies the normalization conditions
\be
\label{71}
\sum_n \;  p(A_n x_n,t) = 1 \; , \qquad 
0 \; \leq \; p(A_n x_n,t)\; \leq \; 1 \; .
\ee

In expression (\ref{70}), it is possible to separate the diagonal part
\be
\label{72}
f(A_n x_n,t) \equiv \sum_\al \; | \; b_{n\al} \; |^2 
\lgl \; \al A_n \; | \; \hat\rho(t) \; | \; A_n\al \; \rgl
\ee
and the nondiagonal part
\be
\label{73}
 q(A_n x_n,t) \equiv \sum_{\al\neq\bt} \;  b^*_{n\al} b_{n\bt}
\lgl \; \al A_n \; | \; \hat\rho(t) \; | \; A_n\bt \; \rgl \; .
\ee
The diagonal part has the meaning of the rational fraction of the total probability 
(\ref{70}), because of which it is called the {\it rational fraction} and is assumed 
to satisfy the normalization condition
\be
\label{74}
 \sum_n \;  f(A_n x_n,t) = 1 \; , \qquad 
0 \; \leq \; f(A_n x_n,t)\; \leq \; 1 \; .
\ee
The rational fraction satisfies the standard properties of classical probabilities.

The nondiagonal part is caused by the quantum interference of emotions and fulfills 
the conditions
\be
\label{75}
\sum_n \;  q(A_n x_n,t) = 0 \; , \qquad 
-1 \;\leq \; q(A_n x_n,t)\; \leq \; 1 \; .
\ee
As far as emotions describe the quality of alternatives, the quantum term 
(\ref{75}) can be called the {\it quality factor}. Being due to quantum 
interference, it also can be named the {\it quantum factor}. And since the 
quality of alternatives characterizes their attractiveness, the term (\ref{75}) 
can be called the {\it attraction factor}.  

Thus the prospect probability (\ref{70}) reads as the sum of the rational fraction 
and the quality factor:
\be
\label{76}
  p(A_n x_n,t) =  f(A_n x_n,t) +  q(A_n x_n,t) \; .
\ee

\section{Conditional behavioral probability}

Sequential choices in quantum decision theory can be treated by analogy with Sec. 3. 
The a priori probability at any time $t < t_0$ is defined in Eq. (\ref{70}), provided 
no explicit choice has been done before the time $t_0$. Just until this time, the 
a priori probability of an alternative $A_n$ is
\be
\label{77}
 p(A_n x_n,t_0-0) = {\rm Tr} \; \hat\rho(t_0-0) \; \hat P(A_n x_n) \; .
\ee
If at the moment of time $t_0$ a choice has been made and an alternative $A_n$ is 
certainly chosen, then the a posteriori probability becomes
\be
\label{78}
p(A_n x_n,t_0+0) = 1\;  .
\ee
This implies the reduction of the probability
\be
\label{79}
p(A_n x_n,t_0-0) \longmapsto p(A_n x_n,t_0+0)
\ee
and the related state reduction
\be
\label{80}
\hat\rho(t_0 - 0) \longmapsto \hat\rho(A_n x_n,t_0+0)  \; .
\ee
Equation (\ref{78}), asserting that 
\be
\label{81}
{\rm Tr} \; \hat\rho(A_n x_n,t_0+0) \; \hat P(A_nx_n) = 1 \; ,
\ee
possesses the solution
\be
\label{82}
\hat\rho(A_n x_n,t_0+0) = \frac{\hat P(A_nx_n)\hat\rho(t_0-0)\hat P(A_nx_n)}
{{\rm Tr}\hat\rho(t_0-0)\hat P(A_nx_n)} \;   .
\ee
The state (\ref{82}) serves as an initial condition for the new dynamics prescribed by
the equation 
\be
\label{83}
\hat\rho(A_n x_n,t) = 
\hat U(t,t_0) \; \hat\rho(A_n x_n,t_0+0) \; \hat U^+(t,t_0) \; .
\ee

The a priori probability of choosing an alternative $B_k$ at any time $t > t_0$, after 
the alternative $A_n$ has certainly been chosen, is the conditional probability
\be
\label{84}
p(B_kx_k,t|A_nx_n,t_0) = {\rm Tr}\; \hat\rho(A_nx_n,t) \; \hat P(B_kx_k) \;  .
\ee
Introducing the joint behavioral probability
\be
\label{85}
p(B_kx_k,t,A_nx_n,t_0) \equiv {\rm Tr}\; \hat U(t,t_0) \; \hat P(A_nx_n) \;
\hat\rho(t_0-0) \; \hat P(A_nx_n) \; \hat U^+(t,t_0) \; \hat P(B_kx_k) \; ,
\ee
allows us to represent the conditional probability in the form
\be
\label{86}
p(B_kx_k,t|A_nx_n,t_0) = \frac{p(B_kx_k,t,A_nx_n,t_0)}{p(A_nx_n,t_0-0)} \;  .
\ee

If the probability of choosing an alternative $B_k$ is evaluated immediately after 
$t_0$, then we need to consider the conditional probability
\be
\label{87}
p(B_kx_k,t_0+0|A_nx_n,t_0) = {\rm Tr}\; \hat\rho(A_nx_n,t_0+0)\; \hat P(B_kx_k)
\ee
and the joint probability
\be
\label{88}
 p(B_kx_k,t_0+0,A_nx_n,t_0) = {\rm Tr}\; \hat P(A_nx_n) \; 
\hat\rho(t_0-0) \; \hat P(A_nx_n) \; \hat P(B_kx_k) \; .
\ee
As a result, the conditional probability (\ref{87}) becomes
\be
\label{89}
 p(B_kx_k,t_0+0|A_nx_n,t_0) = 
\frac{p(B_kx_k,t_0+0,A_nx_n,t_0)}{p(A_nx_n,t_0-0)} \; .
\ee
For the joint probability, we obtain
\be
\label{90}  
 p(B_kx_k,t_0+0,A_nx_n,t_0) = 
|\; \lgl \; x_k B_k\; | \; A_n x_n \; \rgl \; |^2 \; p(A_nx_n,t_0-0) \; .
\ee
Therefore the conditional behavioral probability is
\be
\label{91}
 p(B_kx_k,t_0+0|A_nx_n,t_0) = 
|\; \lgl \; x_k B_k\; | \; A_n x_n \; \rgl \; |^2 \;  .
\ee

\section{Symmetry of behavioral probabilities}

Considering the symmetry properties of behavioral probabilities, it is useful to 
remember that, strictly speaking, emotions are contextual and can vary in time. In 
an approximate picture, it is possible to assume that emotions are mainly associated 
with the corresponding alternatives and are approximately the same at all times. 
Then the symmetry properties of the probabilities can be studied with respect to 
the interchange of the order of the prospects $A_n x_n$ and $B_k x_k$. Keeping 
in mind this kind of the order interchange, we can conclude that the symmetry 
properties of behavioral probabilities are similar to the order symmetry of the 
quantum probabilities examined in Sec. 4. 

Thus for any time $t > t_0$, both the conditional and the joint behavioral 
probabilities are not order symmetric with respect to the prospect interchange,
\be
\label{92}
p(A_nx_n,t|B_kx_k,t_0)  \neq p(B_kx_k,t|A_nx_n,t_0) \qquad ( t > t_0 ) \; ,
\ee
and
\be
\label{93}
p(A_nx_n,t,B_kx_k,t_0)  \neq p(B_kx_k,t,A_nx_n,t_0) \qquad ( t > t_0 ) \; ,
\ee
which is analogous to Eq. (40). 

When the second decision is being made at the time $t = t_0 + 0$ immediately after 
the first choice has been accomplished at the time $t_0$, the conditional behavioral 
probability is order symmetric, 
\be
\label{94}
p(A_nx_n,t_0+0|B_kx_k,t_0) = p(B_kx_k,t_0+0|A_nx_n,t_0) \;  ,
\ee
but the joint probability, generally, is not order symmetric,
\be
\label{95}
 p(A_nx_n,t_0+0,B_kx_k,t_0) \neq p(B_kx_k,t_0+0,A_nx_n,t_0) \;  ,
\ee
which is similar to property (\ref{47}). 

If one makes the immediate sequential choices, and in addition the prospect 
projectors commute with each other, so that
$$
[\; \hat P(A_nx_n) , \; \hat P(B_kx_k) \; ] = 0 \; ,
$$
then the joint behavioral probability becomes order symmetric,
\be
\label{96}
p(A_nx_n,t_0+0,B_kx_k,t_0) = p(B_kx_k,t_0+0,A_nx_n,t_0) \; .
\ee
This property is in agreement with Eq. (\ref{48}). 

Recall that the conditional probability (\ref{93}), because of its form (\ref{91}),  
is symmetric in any case, whether the prospect operators commute or not. 

As is seen, the symmetry properties of the quantum probabilities are in variance 
with the properties (\ref{33}) and (\ref{34}) of the classical probabilities, 
according to which the joint classical probability is order symmetric, while the 
conditional classical probability is not order symmetric.  

The absence of the order symmetry in classical conditional probability is evident 
from the definition (\ref{32}). Empirical investigations \cite{Boyer_24,Boyer_25} 
also show that the conditional probability is not order symmetric. However, as is 
seen from equality (94), the quantum conditional probability is order symmetric. 
Does this mean that the quantum conditional probability cannot be applied to the 
realistic human behavior?

To answer this question, it is necessary to concretize the realistic process of 
making decisions. In reality, any decision is not a momentary action, but it takes 
some finite time. The modern point of view accepted in neurobiology and psychology 
is that the cognition process, through which decisions are generated, involves three 
stages: the process of stimulus encoding through which the internal representation 
is generated, followed by the evaluation of the stimulus signal and then by decoding 
of the internal representation to draw a conclusion about the stimulus that can be 
consciously reported \cite{Woodford_4,Libert_26}. It has been experimentally 
demonstrated that awareness of a sensory event does not appear until the delay time 
up to $0.5$ s after the initial response of the sensory cortex to the arrival of 
the fastest projection to the cerebral cortex \cite{Libert_26,Teichert_27}. About 
the same time is necessary for the process of the internal representation decoding. 
So, the delay time of about $1$ s is the minimal time for the simplest physiological 
processes involved in decision making. Sometimes the evaluation of the stimulus 
signal constitutes the total response time, necessary for formulating a decision, 
of about $10$ s \cite{Hochman_28}. In any case, the delay time of order $1$ s seems 
to be the minimal period of time required for formulating a decision. This assumes 
that in order to consider a sequential choice as following immediately after the 
first one, as is necessary for the quantum conditional probability (\ref{89}) or
(\ref{91}), the second decision has to follow in about $1$ s after the first choice. 
However, to formulate the second task needs time, as well as some time is required 
for the understanding the second choice problem. This process demands several 
minutes. 

In this way, the typical situation in the sequential choices is when the temporal 
interval between the decisions is of the order of minutes, which is much longer 
than the time of $1$ s necessary for taking a decision. Therefore the second choice 
cannot be treated as following immediately after the first one, hence the form of 
the conditional probability (\ref{91}) is not applicable to such a situation. For 
that case, one has to use expression (\ref{86}) which is not order symmetric, in 
agreement with the inequality (\ref{92}) and empirical observations.

Thus the decisions can be considered as following immediately one after the other 
provided the temporal interval between them is of the order of $1$ s. Such a short 
interval between subsequent measurements could be realized in quantum experiments, 
but it is not realizable in human decision making, where the interval between 
subsequent decisions is usually much longer than $1$ s. Hence the form of the 
conditional probability (\ref{91}), that one often calls the L\"{u}ders probability, 
is not applicable to human problems, but expression (\ref{86}), valid for a finite 
time interval between decisions, has to be employed. The latter is not order 
symmetric similarly to the classical conditional probability.       

Concluding, quantum probabilities, whose definition takes into account dynamical 
processes of taking decisions, are more general than simple classical probabilities 
(\ref{32}), hence can be applied to a larger class of realistic human decision problems.

\section*{Acknowledgment}

The author is grateful to J. Harding and H. Nguyen for fruitful discussions and to
E.P. Yukalova for useful advice.

\end{document}